\documentclass[11pt]{article}

\usepackage{acl}

\usepackage{times}
\usepackage{latexsym}
\usepackage[T1]{fontenc}
\usepackage[utf8]{inputenc}
\usepackage{microtype}
\usepackage{inconsolata}

\usepackage{amsmath}
\usepackage{amssymb}
\usepackage{booktabs}
\usepackage{multirow}
\usepackage{array}
\usepackage{graphicx}
\usepackage{adjustbox}   
\usepackage{float}       
\usepackage{tikz}
\usetikzlibrary{positioning, arrows.meta, fit, backgrounds, calc, shapes.geometric}
\usepackage{pifont}       
\usepackage{caption}
\usepackage[inline]{enumitem}
\usepackage{xcolor}
\usepackage{xspace}
\usepackage{xcolor}

\usepackage{xcolor}

\newcommand{\method}{SAB\xspace}

\newcommand{\Btab}{B_{\mathrm{tab}}}
\newcommand{\Embrow}{\mathrm{Emb}_{\mathrm{row}}}
\newcommand{\Embcol}{\mathrm{Emb}_{\mathrm{col}}}
\newcommand{\Embprow}{\mathrm{Emb}_{\mathrm{prow}}}
\newcommand{\Embpcol}{\mathrm{Emb}_{\mathrm{pcol}}}
\newcommand{\IDxrow}{\mathrm{ID}_{\mathrm{row}}}
\newcommand{\IDxcol}{\mathrm{ID}_{\mathrm{col}}}

\title{
DeepTable: Structural Attention Biases and Tree Path Encoding for Hierarchical Table Understanding
}

\author{
Jyun-Ying Yen, Cheng-Kuan Lin, Yu-Chee Tseng \\
Department of Computer Science \\
National Yang Ming Chiao Tung University, Taiwan
}

\begin{document}
\maketitle

\begin{abstract}

Large language models (LLMs) have demonstrated strong performance in table understanding. However, they typically process table content and headers as linearized token sequences. This representation weakens the two-dimensional and hierarchical structural relationships encoded by multi-level row and column headers. 
Existing parameter-efficient fine-tuning methods incorporate basic row and column information but do not explicitly capture the rich structural dependencies induced by hierarchical table headers.
We propose \textbf{DeepTable}, a structure-aware approach for table understanding with LLMs. DeepTable comprises two complementary components. 
Structural Attention Bias (SAB) introduces learnable biases into the
attention logits to explicitly represent whether pairs of table tokens
share the same row or column. 
Tree Path Encoding (TPE) represents each table token using the ancestor paths of its row and column headers, preserving its position within the multi-level table structure. 
We integrate DeepTable with TableLoRA \citep{he2025tablelora} to inject structural information into parameter-efficient adaptation.
Across three LLM backbones, DeepTable consistently improves the corresponding TableLoRA baselines on three table question answering benchmarks, achieving average gains of 7.42 points on HiTab, 3.23 points on WikiTQ, and 2.01 BLEU points on FeTaQA. These results demonstrate the effectiveness of the proposed structural biases across different LLM backbones.
\end{abstract}

\section{Introduction}
\label{sec:intro}

Table understanding aims to answer natural-language queries using both
the content and structure of tables. Unlike natural-language text, tables
organize information through two-dimensional layouts defined by rows,
columns, and headers. Many real-world tables further contain hierarchical
row and column headers, which organize cells into nested groups and define
relationships across different levels of the table. Accurately representing
these relationships is therefore important for table understanding.

Recent approaches to table understanding can be broadly divided into
three directions. The first develops dedicated models through large-scale
pretraining or instruction tuning
\citep{li2024tablegpt,zhang2024tablellm,zhuang2024structlm}. The second
keeps the underlying model frozen and performs reasoning through prompting
or agent-based pipelines \citep{wang2024chainoftable}. The third adopts
parameter-efficient fine-tuning (PEFT), which updates only a small number
of parameters while keeping the pretrained model frozen. Our work follows the third direction and focuses on enhancing structural modeling within parameter-efficient fine-tuning.

A fundamental challenge for PEFT-based table understanding is that tables must first be serialized into one-dimensional token sequences. Although serialization preserves table content in textual form, it does not explicitly encode the two-dimensional structural relationships induced by multi-level row and column headers. Consequently, these structural relationships must be inferred implicitly from the serialized sequence.

Existing PEFT methods generally provide only limited structural modeling.
For example, TableLoRA~\citep{he2025tablelora}  incorporates learnable row and column coordinate embeddings.
While effective, these embeddings represent each token using \textit{flat} two-dimensional coordinates and do not explicitly encode pairwise row or column relationships or the hierarchical header paths that define a token’s position within a multi-level table.
Moreover,
TableLoRA reports smaller improvements on tables with deeper header
hierarchies, motivating the use of additional structural signals for
hierarchical tables.

To address these limitations, we propose \textbf{DeepTable}, a
structure-aware PEFT approach with two complementary components.
\textbf{Structural Attention Bias} (SAB) introduces learnable biases into
the attention logits to explicitly represent whether pairs of table
tokens share the same row or column. \textbf{Tree Path Encoding} (TPE)
represents each token using the ancestor paths of its top and left
headers, thereby encoding information about its position within the
corresponding header hierarchies. Our implementation is built upon the TableLoRA architecture while remaining fully parameter-efficient.

Experiments on three table question answering benchmarks using three LLM backbones demonstrate the effectiveness of DeepTable. Averaged across the three LLM backbones, DeepTable improves HiTab by 7.42 points, WikiTQ by 3.23 points, and FeTaQA by 2.01 BLEU points over the corresponding TableLoRA baselines.


\paragraph{Contributions.}
Our main contributions are threefold.
First, we introduce SAB, which explicitly models pairwise row and column
relationships through learnable structural attention biases.
Second, we introduce TPE, which represents the hierarchical position of
each table token through its top- and left-header ancestor paths.
Third, we integrate SAB and TPE into TableLoRA and demonstrate consistent
improvements over its published results on three table question answering
benchmarks.

\section{Related Work}
\label{sec:related}

We review related work on structure-aware table understanding,
parameter-efficient adaptation for table understanding, and structural
modeling in Transformer architectures.

\paragraph{Structure-aware table understanding.}

Modeling structural information has long been a central problem in table
understanding. TAPAS \citep{herzig2020tapas} augments BERT with row,
column, and rank embeddings. TABBIE \citep{iida2021tabbie} models
row-wise and column-wise dependencies using separate Transformer
encoders, while TUTA \citep{wang2021tuta} introduces tree-based
positional encodings for generally structured tables. These approaches
incorporate table structure through specialized encoder architectures,
whereas DeepTable models structural information within a
parameter-efficient adaptation framework.

\paragraph{Parameter-efficient adaptation for table understanding.}

Recent work has adapted pretrained language models to table understanding
through instruction tuning or task-specific fine-tuning
\citep{li2024tablegpt,zhang2024tablellm,zhuang2024structlm}.
Parameter-efficient fine-tuning methods, including LoRA
\citep{hu2022lora}, prompt tuning \citep{lester2021power}, and p-tuning
\citep{liu2024gpt}, reduce adaptation cost by updating only a
small subset of model parameters. For table understanding, TableLoRA
\citep{he2025tablelora} extends LoRA with learnable row and column
coordinate embeddings. DeepTable builds on this setting by additionally
modeling pairwise row and column relationships and hierarchical header
paths.

\paragraph{Structural modeling in Transformers.}

Structural priors can be incorporated into Transformer models through
positional representations and attention mechanisms. T5
\citep{raffel2020t5} introduces learned relative-position biases,
whereas ALiBi \citep{press2022alibi} employs linear attention biases
based on token distance. For table understanding, TableFormer
\citep{yang2022tableformer} incorporates learnable attention-logit
biases for row, column, and cell relationships. SAT
\citep{zhang2020sat} and MATE \citep{eisenschlos2021mate} instead use
fixed attention patterns derived from table layouts.

Hierarchical representations have also been studied for other structured
inputs. Tree-LSTM \citep{tai2015treelstm} models tree-structured data
through recursive computation, while HiBERT
\citep{zhang2019hibert} learns hierarchical document representations.
DeepTable combines structural attention with header-path encodings in a
parameter-efficient framework for hierarchical table understanding.

\section{Method}
\label{sec:method}

DeepTable enhances hierarchical table understanding by incorporating structural information beyond flat row and column coordinate representations. 
It consists of two complementary components: Structural Attention Bias (SAB), which models structural relationships between table tokens, and Tree Path Encoding (TPE), which encodes the hierarchical organization of table headers.

We first review the underlying TableLoRA adaptation framework (\S\ref{ssec:prelim}), then describe our table serialization and tokenization scheme (\S\ref{ssec:token}), and finally present SAB (\S\ref{ssec:sab}) and TPE (\S\ref{ssec:hier}).


\begin{figure*}[t]
\centering
\IfFileExists{overview.png}{%
  \includegraphics[width=\textwidth]{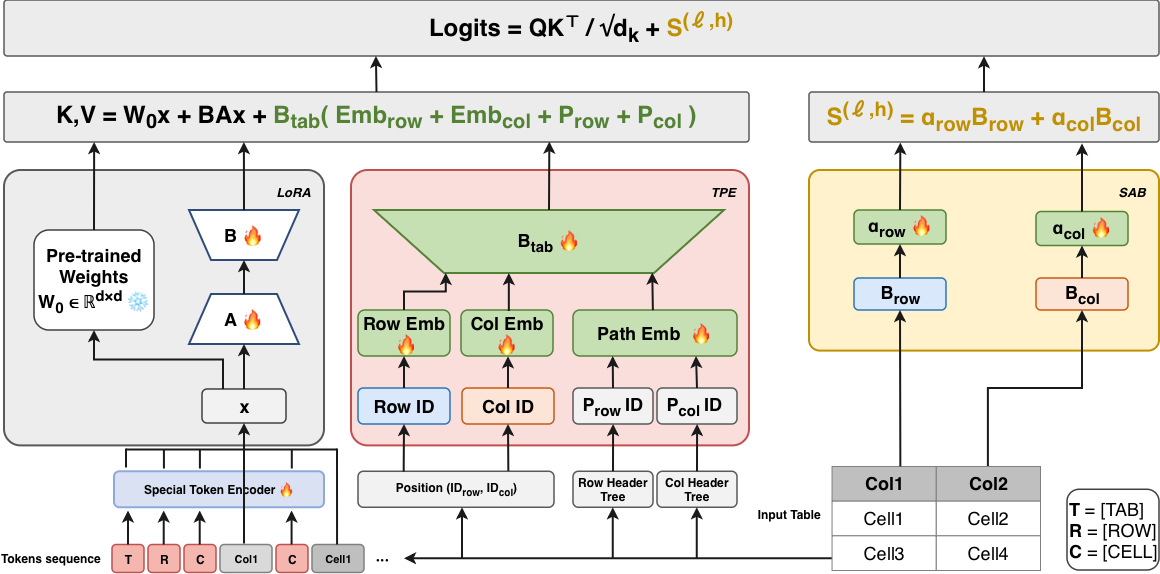}%
}{%
  \fbox{\parbox[c][42mm][c]{0.98\textwidth}{\centering\itshape
  Figure placeholder: export \texttt{overview.drawio} as \texttt{overview.png}
  (File $\to$ Export as $\to$ PNG, 300\,dpi, with \emph{Crop}) and upload it to the project root,
  then recompile.}}%
}
\caption{
Architecture of DeepTable. A hierarchical table is serialized into a token sequence and processed by a pretrained LLM with LoRA adaptation.
DeepTable extends TableLoRA with two structural modules.
Tree Path Encoding (TPE) encodes the row and column paths of each table token, while Structural Attention Bias (SAB) introduces learnable row- and column-aware attention biases into the self-attention mechanism.
}

\label{fig:overview}
\end{figure*}

\subsection{Background: TableLoRA}
\label{ssec:prelim}

TableLoRA augments the key ($K$) and value ($V$) projections with
learnable row and column coordinate embeddings. For each transformer
layer and table token $t$ corresponding to cell
$(\IDxrow,\IDxcol)$, the hidden representation is
\begin{equation}
\label{eq:tablelora}
h = W_0x + BAx + \Btab\,\mathbf{c}_t ,
\end{equation}
where $h,x\in\mathbb{R}^{d}$ denote the output and input hidden states,
$W_0\in\mathbb{R}^{d\times d}$ is the frozen pretrained weight,
$B\in\mathbb{R}^{d\times r}$ and
$A\in\mathbb{R}^{r\times d}$ are the LoRA matrices, and
$\Btab\in\mathbb{R}^{d\times r}$ projects the coordinate embedding into
the hidden space.

The coordinate embedding of token $t$ is
$
\mathbf{c}_t
=
\Embrow(\IDxrow)
+
\Embcol(\IDxcol)$,where
$\Embrow$ and $\Embcol$ are learnable row and column embedding
functions, respectively. For non-table tokens, we set
$\IDxrow=\IDxcol=0$, resulting in zero impact to Eq.~\ref{eq:tablelora}.

The resulting representation is based solely on flat row and column
coordinates. It does not explicitly encode pairwise structural
relations or hierarchical header structures.

\subsection{Table Content Tokenization}
\label{ssec:token}

The input table is serialized into a sequence of tokens.
The entire table begins with a \texttt{[TAB]} token. Each row begins
with a \texttt{[ROW]} token, and each cell begins with a
\texttt{[CELL]} token followed by its textual content. All special
tokens are embedded by a trainable \emph{Special Token Encoder}
(Fig.~\ref{fig:overview}), while ordinary cell-content tokens use the
frozen vocabulary embeddings of the pretrained LLM.

Tables are serialized in row-major order. Header cells and ordinary
cells are treated uniformly as textual tokens. For column headers, each
header level is serialized as a separate row, with consecutive levels
separated by \texttt{[ROW]} tokens. For row headers, the lowest-level
header and its associated data cells occupy the same row, while
ancestor headers are duplicated in that row.

\begin{figure*}[h]
\centering
\includegraphics[width=\textwidth]{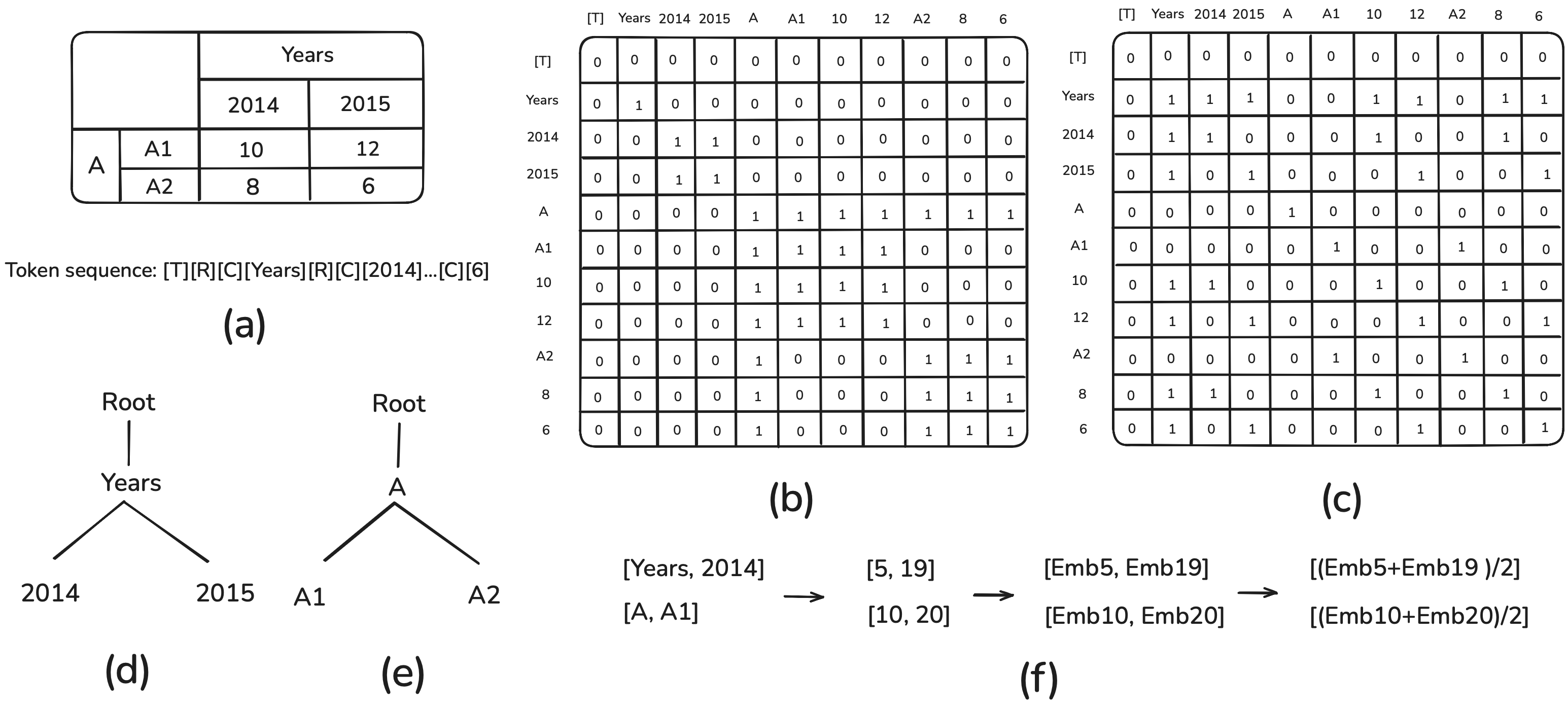}
\caption{
Illustration of the structural representations used in DeepTable.
(a) A hierarchical table and its serialized token sequence;
(b--c) row- and column-adjacency matrices,
$\mathbf{B}_{\mathrm{row}}$ and $\mathbf{B}_{\mathrm{col}}$
(special tokens \texttt{[ROW]} and \texttt{[CELL]} are omitted for clarity);
(d--e) row-header and column-header trees;
(f) examples of row-header and column-header paths and their corresponding embeddings.
}

\label{fig:example}
\end{figure*}

The serialized token sequence is then fed into DeepTable, as illustrated in Fig.~\ref{fig:overview}, which follows the standard LoRA adaptation process.
Equation
\ref{eq:tablelora} is independently applied to the key and value
projections at each transformer layer.
Fig.~\ref{fig:example}(a) shows an example of the serialization process.

\subsection{Structural Attention Bias (SAB)}
\label{ssec:sab}

TableLoRA \citep{he2025tablelora} represents each table token using row and column coordinate
embeddings. While these embeddings encode the position of each token,
they do not explicitly model structural relationships between tokens.
A straightforward approach is to inject predefined row/column biases
into the attention logits. TableLoRA investigated such a variant by
assigning a fixed bias of $+1$ to tokens within the same cell and
$+0.5$ to tokens in the same row or column. However, this strategy
performed worse than the original 2D-LoRA baseline ($40.50$ and $41.00$
vs.\ $48.94$ on HiTab), suggesting that manually designed \textit{fixed} structural
biases are insufficient for modeling hierarchical table structures
\citep{he2025tablelora}.

To address this limitation, SAB introduces \textit{learnable} attention biases.
Instead of using predefined
constants, SAB learns how much additional attention should be assigned
between structurally related tokens. Specifically, the attention
operation becomes
\begin{equation}
\label{eq:sab}
H =
\mathrm{softmax}
\left(
\frac{QK^\top}{\sqrt{d_k}}
+
\mathbf{S}
\right)V ,
\end{equation}
where the structural bias is defined as
$
\mathbf{S}
=
\alpha_{\mathrm{row}}
\mathbf{B}_{\mathrm{row}}
+
\alpha_{\mathrm{col}}
\mathbf{B}_{\mathrm{col}}$,
where
$\alpha_{\mathrm{row}}$ and
$\alpha_{\mathrm{col}}$
are learnable scalar parameters that are independent for each layer
and each attention head.

The binary matrices
$\mathbf{B}_{\mathrm{row}},
\mathbf{B}_{\mathrm{col}}
\in
\{0,1\}^{T\times T}$
indicate whether two table tokens belong to the same row or the same
column, respectively, where $T$ is the total number of tokens.
For any two tokens $x_i$ and $x_j$,
\begin{equation}
\mathbf{B}_{\mathrm{row}}[i,j]
=
\begin{cases}
1, & \IDxrow(i){=}\IDxrow(j) 
\\
& 
\land
\mathbf{1}_{tab}[x_i] \land
\mathbf{1}_{tab}[x_j]
\\
0, & \text{otherwise}.
\end{cases}
\end{equation}
where
$\mathbf{1}_{tab}[\cdot]$
is true iff the token corresponds to a table token
(including both headers and data cells).
Special tokens therefore always receive zero structural bias.
The definition of
$\mathbf{B}_{\mathrm{col}}$
is analogous.

Figs.~\ref{fig:example}(b)–(c) illustrate the row- and column-adjacency matrices $\mathbf{B}_{\mathrm{row}}$ and $\mathbf{B}_{\mathrm{col}}$.
For example,
\texttt{2014} and \texttt{2015} belong to the same header row,
while
\texttt{2014}
shares a column with
\texttt{10}
and
\texttt{8}.
In contrast,
\texttt{2014}
and
\texttt{12}
share neither a row nor a column.
The parent headers further demonstrate that hierarchical structures
naturally propagate row and column membership to their descendants.
For instance,
\texttt{years}
shares a column with every cell beneath
\texttt{2014}
and
\texttt{2015},
while
\texttt{A}
shares a row with every cell beneath
\texttt{A1}
and
\texttt{A2}.

Overall, SAB explicitly encourages attention between tokens belonging
to the same row or column, allowing structural relationships to be
modeled directly within the attention mechanism. Removing
$\mathbf{S}$
reduces DeepTable to the original TableLoRA.

\subsection{Tree Path Encoding (TPE)}
\label{ssec:hier}

SAB explicitly models same-row and same-column relationships through
attention biases. However, it does not explicitly encode the hierarchical
organization of table headers. For example, in
Fig.~\ref{fig:example}(b), the matrix
$\mathbf{B}_{\mathrm{col}}$ only implicitly indicates that the headers
\texttt{2014} and \texttt{2015} share the same parent,
\texttt{years}. Moreover, a table may contain multiple subheaders with
the same name (e.g., two \texttt{2015} subheaders), making the hierarchy ambiguous.
To address this limitation, TPE explicitly encodes tree paths.

TPE first represents the table headers as \textit{row-header} and \textit{column-header trees}, as illustrated in Fig.~\ref{fig:example}(d)–(e).
For each table token $t$, two
tree paths are defined. 
The \textit{row path} of $t$, denoted by
$\mathrm{row}_t$, consists of the sequence of the nodes from the root to the bottom header node directly above $t$ in the row-header tree. Similarly, the
\textit{column path} of $t$, $\mathrm{col}_t$, is defined in the column-header
tree. For example,
$\mathrm{row}_{\texttt{10}}
=
(\texttt{A}\rightarrow\texttt{A1})$
and
$\mathrm{col}_{\texttt{10}}
=
(\texttt{years}\rightarrow\texttt{2014})$.

These tree paths are then encoded into embedding vectors, using dedicated path-embedding tables $\Embprow$ and $\Embpcol$.
Each header node in either tree is assigned a unique integer identifier, which is then mapped to a trainable embedding vector through an embedding table.
The embedding of the row path of $t$ is computed as the
average of the embeddings of all nodes along the path $
\mathbf{P}_{\mathrm{row}}(t)
=
\frac{1}{|\mathrm{row}_t|}
\sum_{n\in\mathrm{row}_t}
\Embprow(n)$.
The embedding of the column path of $t$,
$\mathbf{P}_{\mathrm{col}}(t)$,
is defined analogously using a
separate embedding table $\Embpcol$. Both embedding tables are
layer-specific and zero-initialized.

Finally, the row-path and column-path embeddings are added to the original row and column coordinate embeddings, and the resulting structural embeddings are incorporated into the LoRA adaptation,
\begin{equation*}
\label{eq:hier}
h
=
W_0 x
+
BAx
+
\Btab
\bigl(
\mathbf{c}_t
+
\mathbf{P}_{\mathrm{row}}(t)
+
\mathbf{P}_{\mathrm{col}}(t)
\bigr).
\end{equation*}
Consequently, each table token is represented by four structural
signals: the row coordinate, the column coordinate, the row-path
embedding, and the column-path embedding.


\section{Experiments}
\label{sec:exp}

\subsection{Experimental Setup}
\label{ssec:setup}
\paragraph{Models and training.}

We use DeepSeek-LLM-7B-Chat~\citep{deepseek2024llm} as our primary
base model. Following TableLoRA~\citep{he2025tablelora}, which also evaluates
Llama-3-8B-Instruct~\citep{llama3herd}, we additionally report results on
this backbone. To further evaluate whether DeepTable generalizes across
different LLM architectures, we conduct the same experiments on
Qwen2.5-7B-Instruct~\citep{qwen2024qwen25}, which was not evaluated in the
original paper. Both Llama-3-8B-Instruct and Qwen2.5-7B-Instruct adopt
grouped-query attention instead of the standard multi-head attention used by
DeepSeek. Unless otherwise specified, we follow the experimental settings of
TableLoRA~\citep{he2025tablelora}.

All models are trained using the same training hyperparameters.
Detailed training settings, including the LoRA rank, learning rate,
batch size, number of epochs, and hardware configuration, are provided
in Appendix~\ref{app:hparams}.

The learning-rate multiplier for SAB and TPE is fixed to a single shared value across all experiments, selected using a grid search on HiTab with DeepSeek
(Appendix~\ref{app:sweep}) and applied unchanged to Qwen2.5.
We report the average performance over four random seeds using greedy
decoding throughout. Additional comparisons with sampling decoding are
provided in Appendix~\ref{app:decoding}.

\begin{table}[t]
\centering\small
\setlength{\tabcolsep}{4pt}
\adjustbox{max width=\columnwidth}{%
\begin{tabular}{lcccc}
\toprule
 & \textbf{HiTab} & \textbf{WikiTQ} & \textbf{FeTaQA} & \textbf{TabFact} \\
\midrule
Structure        & hier. & flat & flat$+$span & flat \\
Multi-lvl col hdr & 78\% & 0\% & 0\% & 0\% \\
Multi-lvl row hdr & 86\% & 0\% & 0\% & 0\% \\
Test size         & 1{,}584 & 4{,}344 & 2{,}003 & 12{,}779 \\
Output            & value & span & sentence & bit \\
Metric            & acc & acc & BLEU & acc \\
\bottomrule
\end{tabular}
}
\caption{
Summary of the tabular datasets.
HiTab is the only dataset with hierarchical headers and therefore serves as our primary benchmark.
FeTaQA is the only generative benchmark, whereas TabFact is the only binary classification benchmark.
}
\label{tab:datasets}
\end{table}

\paragraph{Datasets and metrics.}
We evaluate on the four benchmark datasets used by
TableLoRA~\citep{he2025tablelora}: HiTab~\citep{cheng2022hitab},
WikiTQ~\citep{pasupat2015wikitq}, FeTaQA~\citep{nan2022fetaqa}, and
TabFact~\citep{chen2020tabfact}. Table~\ref{tab:datasets} summarizes
their structural characteristics and output formats.

HiTab evaluates hierarchical-table question answering, WikiTQ
evaluates flat-table question answering, FeTaQA evaluates free-form
table question answering, and TabFact evaluates table fact
verification. 
We follow the standard evaluation metrics: accuracy for
HiTab, WikiTQ, and TabFact, and
BLEU~\citep{post2018sacrebleu} for FeTaQA.

\begin{table*}[t]
\centering
\small
\setlength{\tabcolsep}{5pt}
\begin{tabular}{llcc r@{~}l r@{~}l r@{~}l r@{~}l}
\toprule
{\bf Backbone} & {\bf Method} &{\bf SAB} & {\bf TPE} & \multicolumn{2}{c}{\textbf{HiTab} (acc)$\uparrow$} & \multicolumn{2}{c}{\textbf{WikiTQ} (acc)$\uparrow$}
 & \multicolumn{2}{c}{\textbf{FeTaQA} (BLEU)$\uparrow$} & \multicolumn{2}{c}{\textbf{TabFact} (acc)$\uparrow$} \\
\midrule

& TableLoRA & $-$ & $-$ & 46.94 & & 40.42 & & 27.29 & & {\bf 77.05} & \\
DeepSeek- & DeepTable & \checkmark & $\times$ & {\bf  52.05} & ($+$5.11) & {\underline{42.79}} & ($+$2.37) & 29.52 & ($+$2.23) & {\underline{76.35}} & ($-$0.70) \\
LLM-7B-Chat & DeepTable & $\times$ & \checkmark & 45.83 & ($-$1.11) & 42.30 & ($+$1.88) & {\bf 30.32} & ($+$3.03) & 74.44 & ($-$2.61) \\
& DeepTable& \checkmark & \checkmark & {\underline{ 51.20}} & ($+$4.26) & {\bf 43.08} & ($+$2.66) & {\underline{ 30.27}} & ($+$2.98) & 75.03 & ($-$2.02) \\

\midrule

& TableLoRA & $-$ & $-$& 58.56 & & 53.45 & & 30.23 & & 84.01 & \\
Llama-3-& DeepTable& \checkmark & $\times$ & {\bf 74.01} & ($+15.45$) & {\bf 59.81} & ($+6.36$) & 32.63 & ($+2.40$) & {\bf 85.06} & ($+1.05$) \\
8B-Instruct & DeepTable & $\times$ & \checkmark & 71.83 & ($+13.27$) & 59.19 & ($+5.74$) & {\underline{32.66}} & ($+2.43$) & {\underline{84.40}} & ($+0.39$) \\
& DeepTable& \checkmark & \checkmark & {\underline{73.39}} & ($+14.83$) & {\underline{59.45}} & ($+6.00$) & {\bf 32.79} & ($+2.56$) & 84.36 & ($+0.35$) \\
\midrule
\multirow{4}{*}{Qwen2.5-7B} 
& TableLoRA & $-$ & $-$ & 62.38 & & 52.95 & & 30.49 & & {\underline{81.20}} & \\
& DeepTable & \checkmark & $\times$ & {\underline{65.36}} & ($+$2.98) & {\underline{53.79}} & ($+$0.84) & {\underline{30.96}} & ($+$0.47) & {\bf 81.33} & ($+$0.13) \\
& DeepTable & $\times$ & \checkmark & 62.34 & ($-$0.04) & 53.21 & ($+$0.26) & 30.77 & ($+0.28$) & 80.71 & ($-0.49$) \\
& DeepTable & \checkmark & \checkmark & {\bf 65.56} & ($+$3.18) & {\bf 53.99} & ($+$1.04) & {\bf 30.97} & ($+$0.48) & 81.14 & ($-$0.06) \\
\bottomrule
\end{tabular}

\caption{
Comparison of DeepTable with TableLoRA across four benchmarks using three LLM backbones. Results are reported as four-seed greedy means.
For DeepSeek-LLM-7B-Chat and Llama-3-8B-Instruct, TableLoRA results are taken from~\citep{he2025tablelora}. For Qwen2.5-7B, the TableLoRA baseline is reproduced using the official implementation because this backbone was not evaluated in the original paper. Values in parentheses denote improvements over the corresponding TableLoRA baseline. 
Bold indicates the best result, and underline indicates the second-best result.
}
\label{tab:main}
\end{table*}

The four datasets differ in both table structure and output format.
HiTab is the only benchmark containing genuinely hierarchical headers,
where $78\%$ of tables have multi-level column headers and $86\%$ have
multi-level row headers (up to depth~3). 
Therefore, HiTab serves as our primary benchmarking dataset.
In contrast, WikiTQ and
TabFact contain only flat tables, while FeTaQA contains flat tables
with horizontally spanning column groups ($23\%$ of tables). The output
formats also vary across datasets: HiTab and WikiTQ predict short cell
values, TabFact predicts a binary \texttt{True}/\texttt{False} label,
and FeTaQA generates a free-form sentence (median length 19 words).

\begin{table}[t]
\centering
\setlength{\tabcolsep}{4pt}
\small
\begin{tabular}{cccccc}
\toprule
{\bf SAB} & {\bf TPE}  & {\bf HiTab} & {\bf WikiTQ} & {\bf FeTaQA} & {\bf TabFact} \\
\midrule
 \checkmark & $\times$   & $+$7.85 & $+$3.19 & $+$1.70 & $+$0.16\\
 $\times$ & \checkmark   & $+$4.04 & $+$2.63 & $+$1.91 & $-$0.90 \\
 \checkmark & \checkmark         & $+$7.42 & $+$3.23 & $+$2.01 & $-$0.58 \\
\bottomrule
\end{tabular}
\caption{Average improvements (\%) of DeepTable variants over the corresponding TableLoRA baselines, averaged across the three LLM backbones.}
\label{tab:avg_improvement}
\end{table}

\subsection{Comparison with TableLoRA}
\label{ssec:main}

Table~\ref{tab:main} compares DeepTable with the corresponding
TableLoRA baselines across four table question answering benchmarks and three base models. 
Overall, the proposed structural biases consistently
improve performance on HiTab, WikiTQ, and FeTaQA while maintaining
performance comparable to TableLoRA on TabFact, which contains only flat tables.

These results lead to the following hypotheses.
First, the largest improvements are consistently observed on HiTab, the only benchmark containing hierarchical row and column headers.
This suggests that the proposed structural biases are particularly effective for hierarchical tables (\S\ref{sec:depth}).
Second, the relative contributions of the two proposed biases are task-dependent (\S\ref{sec:mechanism}).
While \method\ consistently provides the largest improvements on HiTab, TPE achieves its largest gains on the generative benchmark FeTaQA.

Across all three base models, the same qualitative trend is observed.
Both \method\ and TPE consistently improve or maintain the
corresponding TableLoRA baselines on the three table question answering
benchmarks, and their combination remains competitive across all four
datasets. The magnitude of improvement, however, varies across
backbones. DeepSeek exhibits the clearest gains, Qwen2.5-7B shows
smaller improvements because its TableLoRA baseline is already
substantially stronger (e.g., $62.38$ vs.\ $46.94$ on HiTab), and
Llama-3-8B-Instruct demonstrates that the proposed structural biases
generalize to a third independently trained backbone. 
Backbone-specific differences are further discussed in
\S\ref{sec:mechanism}.

Table~\ref{tab:avg_improvement} summarizes the average improvements achieved by DeepTable across the three LLM backbones.
Overall, all three variants improve performance over the corresponding TableLoRA baselines. While DeepTable achieves the largest average improvements on WikiTQ and FeTaQA, DeepTable w/o TPE performs better on HiTab and TabFact, suggesting that SAB and TPE provide complementary structural benefits whose relative effectiveness varies across datasets.

HiTab additionally annotates each query with its aggregation
function~\citep{cheng2022hitab}. We therefore stratify the $4$-seed
paired accuracy by aggregation type in Table~\ref{tab:agg} to understand
where the proposed structural biases are most effective. This
fine-grained analysis is currently available only for DeepSeek-LLM-7B;
we leave the corresponding Qwen2.5-7B and Llama-3-8B analyses for future
work.

Table~\ref{tab:agg} reveals three notable patterns. First, the largest
single \method\ gain is observed on \emph{none} queries
($+9.2$~pp, $n=1133$), which require retrieving a single table cell.
Second, comparison and ranking queries (pair-argmax, argmax, opposite,
and pair-argmin) also benefit from \method, although the improvements
are more modest ($+1$--$4$~pp). Finally, arithmetic queries (division
and summation) remain unsolved, with all configurations, including
TableLoRA, achieving $0\%$ accuracy.

\begin{table*}[t]
\centering
\small
\begin{tabular}{llccccc}
\toprule
\multirow{2}{*}[-0.5ex]{{\bf Category}}& \multirow{2}{*}[-0.5ex]{{\bf Aggregation}} & \multirow{2}{*}[-0.5ex]{$\mathbf{n}$} & \multirow{2}{*}[-0.5ex]{{\bf TableLoRA}} &\multicolumn{3}{c}{{\bf DeepTable}}\\
\cmidrule(lr){5-7}
& & &
& {\bf w/o TPE} & {\bf w/o SAB}  & {\bf Full} \\
\midrule
Lookup &
 none      & 1133 & 53.1 & {\bf 62.1}~($+$9.2)  & 54.4~($+$1.3)  & 61.3~($+$8.2)  \\
\midrule
\multirow{4}{*}{Comparison/Ranking} &
pair-argmax  &  93  & 38.7 & 39.8~($+$1.1)  & 36.0~($-$2.7)  & {\bf 41.4}~($+$2.7) \\
 & argmax       &  56  & 30.4 & {\bf 33.1}~($+$2.7)  & 29.1~($-$1.3)  & 30.4~($+$0.0)  \\
& opposite     &  50  & 49.0 & {\bf 53.0}~($+$4.0)  & 51.0~($+$2.0)  & 51.5~($+$2.5)  \\
& pair-argmin  &  21  & 44.0 & {\bf 45.2}~($+$1.2)  & 39.2~($-$4.8)  & 44.0~($+$0.0)  \\
\midrule
\multirow{3}{*}{Arithmetic}  & diff  &  56 &  5.4 & 7.2~($+$1.8)  & 7.2~($+$1.8)  & {\bf 8.1}~($+$2.7)  \\
& div   &  76 &  0.0 & 0.0~($+$0.0)           & 0.0~($+$0.0)           & 0.0~($+$0.0)           \\
& sum   &  22 &  0.0 & 0.0~($+$0.0)           & 0.0~($+$0.0)           & 0.0~($+$0.0)           \\
\bottomrule
\end{tabular}
\caption{
Performance by aggregation type on HiTab with DeepSeek-LLM-7B-Chat.
Four-seed accuracy (\%) is reported for each aggregation type, with
improvements over TableLoRA shown in parentheses.
Only aggregation types with $n \ge 20$ test instances per seed are shown.
DeepTable achieves its largest gains on lookup and comparison/ranking
queries, whereas arithmetic queries remain difficult for all methods.
}
\label{tab:agg}
\end{table*}

\subsection{Header-Depth Analysis}
\label{sec:depth}
\setlength{\tabcolsep}{5pt}
\begin{table*}[t]
\centering 
\small
\begin{tabular}{llcc r@{~}l r@{~}l r@{~}l c}
\toprule

{\bf Backbone} & {\bf Method} &{\bf SAB} & {\bf TPE} & \multicolumn{2}{c}{{\bf Depth~1}~{\scriptsize($n=353$)}}
 & \multicolumn{2}{c}{{\bf Depth~2~{\scriptsize($n=925$)}}} & \multicolumn{2}{c}{{\bf Depth~3}~{\scriptsize($n=306$)}} & {\bf Monotonic}\\

\midrule

&TableLoRA     & $-$ & $-$ & 42.56 & & 50.38 & &31.94 & & --- \\
DeepSeek- &DeepTable  & \checkmark & $\times$       & \bf{48.93} & ($+$6.37) & {\bf 57.41} & ($+$7.03) & \bf{39.46} & ($+$7.52) & \checkmark \\
LLM-7B-Chat &DeepTable & $\times$ & \checkmark          & 42.70 &  ($+$0.14) & 51.49 & ($+$1.11) & 32.35 & ($+$0.41) & Flat ($\approx$ 0) \\
 &DeepTable    & \checkmark & \checkmark   & 47.66 & ($+$5.10) & 56.57 & ($+$6.19) & 39.05 & ($+$7.11) & \checkmark \\
\midrule
\multirow{4}{*}{Qwen2.5-7B}
&TableLoRA    & $-$ & $-$    & 56.73 & & 68.11 & & 51.55 & & --- \\
&DeepTable  & \checkmark & $\times$       & 58.00 & ($+$1.27) & 70.65 & ($+$2.54) & \bf{57.84} & ($+$6.29) & \checkmark \\
&DeepTable  & $\times$ & \checkmark          & 57.01 & ($+$0.28) & 67.52 & ($-$0.59) & 52.86 & ($+$1.31) & Flat ($\approx$ 0) \\
&DeepTable  & \checkmark & \checkmark     & \bf{59.14} & ($+$2.41) & \bf{ 71.08} & ($+$2.90) & 56.29 & ($+$4.74) & \checkmark \\
\bottomrule
\end{tabular}

\caption{\method's improvement increases with header depth, whereas TPE remains
nearly flat. HiTab four-seed accuracy (\%) and paired $\Delta$ relative
to the TableLoRA baseline are stratified by top-header depth.
The largest gain is consistently observed on depth-3 tables.
A finer left$\times$top depth analysis for DeepSeek is provided in
Appendix~\ref{app:depthgrid}.}
\label{tab:depth}
\end{table*}

To analyze the HiTab results, we further stratify the test set by
top-header depth and compute the $4$-seed paired lift over the
TableLoRA baseline. Table~\ref{tab:depth}
reveals two clear observations. First, \method's lift increases
monotonically with header depth, from $+6.37$ to $+7.03$ and
$+7.52$. Second, TPE's lift remains nearly flat across the three
depth levels ($+0.14$, $+1.11$, and $+0.41$).

The lower block of Table~\ref{tab:depth} repeats the same analysis on
Qwen2.5-7B using the same depth metadata and evaluation protocol. The
same two observations are reproduced. \method's lift again increases
monotonically with header depth ($+1.27\!\to\!+2.54\!\to\!+6.29$),
whereas TPE's lift remains small and shows no monotonic trend
($+0.28$, $-0.59$, and $+1.31$). The overall improvement of \method\
is also statistically significant under a paired $t$-test
($+2.98$~pp, $4/4$ seeds positive, $p\!\approx\!0.04$).

\subsection{Two-Axis Analysis: Structure vs. Hierarchy Complexity}
\label{sec:mechanism}

A table consists of headers and data cells. While TPE explicitly models the
hierarchical organization of table headers, SAB captures the structural
relationships among data cells. These two structural biases are not always
complementary. On some datasets, combining SAB and TPE achieves the best
performance, whereas on others, SAB alone performs better. 
We therefore investigate this issue in detail.

\paragraph{Table Structural Complexity.}
The benefit of \method\ is governed by the structural complexity of the
table, including both table size and header depth. By encouraging
same-row and same-column attention, \method\ directly facilitates the
structural alignment required for locating and comparing table cells.

This explanation is consistent with the aggregation analysis in
Table~\ref{tab:agg}. The largest improvement is observed on
\emph{none} queries, which require retrieving a single target cell,
exactly the type of structural alignment that \method\ is designed to
facilitate. Comparison and ranking queries also benefit from
\method, although the improvements are smaller because alignment is
performed over multiple candidate cells rather than a single target
cell. In contrast, arithmetic queries remain at $0\%$ across all
configurations, indicating that structural bias alone cannot compensate
for numerical reasoning abilities absent from the underlying LLM.

The same explanation also accounts for the depth analysis in Table~\ref{tab:depth}. As header depth increases, a header cell's identity becomes increasingly dispersed across its descendant cells, making structural alignment progressively more difficult. Accordingly, the performance gains of SAB increase monotonically with header depth on both DeepSeek and Qwen2.5-7B.

\paragraph{Generation vs.\ Extraction Tasks.}
The contribution of TPE is governed by the output modality.
TPE enriches each cell with hierarchical header semantics, which are particularly useful for \textit{generation tasks}, where the model must explicitly verbalize header information rather than simply extract cell values.

The above hypothesis is supported by both the aggregation and depth analyses.
Across aggregation types and header depths, TPE provides only marginal or even slightly negative gains on \textit{extraction tasks}, where the goal is to retrieve a table cell rather than generate a sentence.
In these cases, the hierarchical header signal is largely redundant.
We suspect that it sometimes interferes with the two-dimensional coordinate representations. 
In contrast,
TPE achieves its largest improvement on the generative benchmark
FeTaQA, where header semantics naturally become part of the generated
output.

Together, these two properties explain the results in Table~\ref{tab:main}. HiTab combines high structural complexity with
cell extraction, making \method\ the dominant contributor. FeTaQA
requires sentence generation, where TPE achieves its largest
improvement. WikiTQ benefits from combining both structural biases.

\paragraph{Control Experiment on TabFact.}
TabFact is the only benchmark that is both structurally simple and
binary (a single \texttt{True}/\texttt{False} output). Accordingly,
\method\ remains essentially on par with the TableLoRA baseline, while
TPE provides little benefit. The same trend is also observed on
Qwen2.5-7B (Appendix~\ref{app:tabfact_tf}), where both effects shrink
toward zero rather than reversing.

\paragraph{Cross-Backbone Generalization.}
The above findings remain consistent across both DeepSeek and Qwen2.5-7B.
The primary difference is not
whether the two biases help, but how much they contribute once
combined. On HiTab, the full model slightly outperforms \method\ alone
on Qwen2.5-7B, whereas \method\ alone performs better on DeepSeek.
Similarly, on FeTaQA, TPE provides a substantially larger standalone
gain on DeepSeek than on Qwen2.5-7B. We attribute this difference to
the stronger instruction-following capability of Qwen2.5-7B, which
likely already captures part of the hierarchical header semantics that
TPE explicitly injects. 
Consequently, the effectiveness of TPE appears to depend more strongly on the underlying backbone, whereas the benefits of SAB are consistently observed across both models.

\section{Conclusion}
\label{sec:conclusion}

DeepTable introduces two structural biases for hierarchical table
understanding. SAB adds learnable attention biases for same-row and
same-column cell pairs at the per-layer, per-attention-head level,
whereas TPE injects hierarchical header-path information into each cell
representation.
On DeepSeek-LLM-7B-Chat, the combined model outperforms TableLoRA on all three table question answering benchmarks (HiTab: $+5.11$, WikiTQ: $+2.66$, and FeTaQA: $+3.03$ BLEU), while maintaining comparable performance on TabFact.

Our analyses further distinguish the empirical behavior of the two
biases. \method\ provides substantial gains on lookup and
comparison/ranking queries, and its improvement increases with header
depth on both DeepSeek-LLM-7B-Chat and Qwen2.5-7B. In contrast,
TPE provides limited or inconsistent gains across most aggregation
types and header depths, but contributes more prominently on the
free-form generation benchmark. The overall comparison across
DeepSeek-LLM-7B-Chat, Llama-3-8B-Instruct, and Qwen2.5-7B also shows
that the relative effectiveness of the two biases can vary across
backbones and datasets.

Future work may explore adaptive mechanisms that select structural biases according to the task.



\section*{Limitations}

\paragraph{Backbone coverage.}
We evaluate DeepTable on three open-source LLMs
(DeepSeek-LLM-7B-Chat, Qwen2.5-7B-Instruct, and
Llama-3-8B-Instruct), all in the 7--8B parameter range.
Whether the same improvements generalize to substantially larger
or closed-source LLMs remains to be investigated.
Moreover, the effectiveness of explicit structural biases may decrease
as stronger base models implicitly capture richer table structures.

\paragraph{Evaluation scope.}
Our fine-grained analyses that explain \emph{why} DeepTable improves
performance include the aggregation-type breakdown
(Table~\ref{tab:agg}, DeepSeek-LLM-7B-Chat only), the header-depth analysis
(Table~\ref{tab:depth} and Appendix~\ref{app:depthgrid}), and the TabFact True/False breakdown (Appendix~\ref{app:tabfact_tf}); the latter two are conducted on both DeepSeek-LLM-7B-Chat and Qwen2.5-7B-Instruct.
For Llama-3-8B-Instruct, we report only the overall results in
Table~\ref{tab:main}.
In addition, the magnitude of DeepTable's improvements depends on the decoding strategy (Appendix~\ref{app:decoding}).
We report greedy decoding throughout for consistency, but the absolute performance and improvement over TableLoRA may differ under other generation settings.

\paragraph{Arithmetic reasoning remains challenging.}
On HiTab queries that require arithmetic operations, such as division and summation, all evaluated methods, including the TableLoRA baseline and DeepTable, achieve 0\% accuracy (Table~\ref{tab:agg}).
While the proposed structural biases help identify and align relevant table cells, they do not improve the numerical reasoning capability of the underlying LLM.
Consequently, DeepTable's performance on arithmetic queries remains bounded by the reasoning ability of the base model.

\paragraph{Hyperparameter tuning.}
The learning-rate multiplier for SAB and TPE
($\lambda=1000$) is selected using a sweep on a single HiTab seed with DeepSeek (Appendix~\ref{app:sweep}) and then reused across all datasets, random seeds, and backbone LLMs.
Although this demonstrates that a single setting generalizes well across different conditions, further per-model or per-dataset tuning may yield additional improvements.

\bibliography{refs}

\clearpage
\appendix

\section{Decoding-Protocol Audit}
\label{app:decoding}
During development, we found that evaluation results depend on the decoding protocol.
The base chat model provides a default generation configuration with stochastic sampling
(\texttt{do\_sample=true}, temperature $=0.7$, top-$p=0.95$), which is inherited by the fine-tuning framework unless explicitly overridden.
Under stochastic sampling, both the absolute scores and the observed performance gap differ from those obtained using greedy decoding.
Therefore, all results reported in this paper are evaluated using \emph{greedy decoding}.

Table~\ref{tab:greedy} compares the two decoding protocols on HiTab with DeepSeek-LLM-7B-Chat.
Under stochastic sampling, DeepTable improves over TableLoRA by 2.49 percentage points, whereas the improvement is 1.50 percentage points under greedy decoding.

\section{Learning-Rate Multiplier Sweep}
\label{app:sweep}

The \method/TPE parameter groups are substantially smaller than the base LoRA matrices and therefore require separate learning-rate multipliers.
To determine appropriate values, we performed a two-dimensional grid search over
$(\lambda_{\method},\lambda_{TPE})$ on HiTab using a single random seed (seed~0) under greedy decoding.
Table~\ref{tab:sweep} summarizes the evaluated settings and their corresponding accuracies.
Among the tested combinations, $(1000,1000)$ achieved the highest accuracy.
Accordingly, we use $\lambda_{\method}=\lambda_{TPE}=1000$ in all reported experiments.

\section{Subspace-Overlap and Tree Path Encoding Variant Ablations}
\label{app:ablation}

\paragraph{Subspace overlap between Tree Path Encoding and 2D LoRA.}
To examine the relationship between TPE and the row/column embeddings used in
2D LoRA, we disable the row/column embeddings while retaining TPE (and
\method). On HiTab, removing the row/column embeddings decreases the accuracy
from 45.71\% to 43.75\%. This result shows that the header-path encoding remains
effective even without the row/column embeddings.

\paragraph{Alternative Tree Path Encoding variants.}
We also evaluated several implementation variants of TPE, including replacing
mean pooling with sum pooling, varying the learning-rate multiplier
$\lambda_{TPE}$, and using only the top-header path. None of these variants
consistently outperformed the adopted configuration, and all achieved
performance close to the TableLoRA baseline on HiTab.

Under the current design, TPE shares the projection matrix $\Btab$ with
2D LoRA, making it impossible to increase the capacity of the path embeddings
without modifying the underlying architecture. Exploring decoupled projection
matrices is left for future work.

\begin{table}[t]
\centering
\small
\setlength{\tabcolsep}{4pt}
\begin{tabular}{lrr}
\toprule
{\bf Method} & {\bf Sampling} & {\bf Greedy} \\
\midrule
TableLoRA                       & 40.95 & 45.08  \\
DeepTable   &\textbf{43.45}~(+2.49) & \textbf{46.58}~(+1.50) \\
\midrule
Paired $t$-test  & $p =.015$& $p~{\approx}.06$ \\
\bottomrule
\end{tabular}

\vspace{2pt}
\begin{flushleft}
\footnotesize
\emph{Note.} Results are averaged over four random seeds.
$p$-values are computed using paired $t$-tests.
\end{flushleft}
\caption{Decoding-protocol audit on HiTab with DeepSeek-LLM-7B-Chat. Greedy decoding lowers absolute scores and reduces the performance gap
relative to the default sampling decoder. TableLoRA is our own reproduction under each decoding protocol, and therefore differs slightly from the published TableLoRA number used as the baseline in Table~\ref{tab:main}.
}
\label{tab:greedy}
\end{table}

\begin{table}[t]
\centering
\small
\begin{tabular}{lrr}
\toprule
${\mathbf{(\lambda_{\method}, \, \lambda_{TPE})}}$& {\bf Accuracy} \\
\midrule
$(30,30)$       & 44.95~(-3.41) \\
$(100,1000)$    & 47.10~(-1.26) \\
$(300,100)$     & 47.41~(-0.95) \\
$(300,300)$     & 47.79~(-0.57) \\
$(100,100)$     & 48.74~(+0.38) \\
$(1000,100)$    & 49.49~(+1.13) \\
$(100,300)$     & 50.06~(+1.70) \\
$(2000,2000)$   & 50.63~(+2.27) \\
$(2500,2500)$   & 52.59~(+4.23) \\
$\mathbf{(1000,1000)}$ & {\bf 53.28}~(+4.92) \\
\bottomrule
\end{tabular}
\caption{Learning-rate multiplier sweep on HiTab (seed~0, greedy decoding).
Values in parentheses denote the improvement over the TableLoRA baseline (48.36\%).
}
\label{tab:sweep}
\end{table}

\section{Per-Seed Results and Significance}
\label{app:perseed}
Tables~\ref{tab:app-hitab}--\ref{tab:app-tabfact} report the per-seed results
underlying the averaged results in Table~\ref{tab:main}. For each benchmark, we
present the results for all four random seeds together with the corresponding
average performance. Results are provided for all three backbones
(DeepSeek-LLM-7B-Chat, Llama-3-8B-Instruct, and Qwen2.5-7B-Instruct) under the
same experimental settings used in the main paper, providing a complete
breakdown of the reported results.

\begin{table*}[t]
\centering\small
\setlength{\tabcolsep}{4pt}
\begin{tabular}{llcccccccc}
\toprule
 {\bf Backbone} & {\bf Dataset} & {\bf SAB} & {\bf TPE} & {\bf s0} & {\bf s1} & {\bf s2} & {\bf s3} & {\bf Avg.} \\
\midrule
\multirow{9}{*}{DeepSeek-LLM-7B-Chat}
& \multirow{3}{*}{HiTab} & \checkmark & $\times$ & {\bf 53.79} & {\bf 51.58} & {\bf 51.20} & {\bf 51.64} & {\bf 52.05} \\
&  & $\times$ & \checkmark & 47.16 & 43.18 & 46.15 & 46.84 & 45.83 \\
&  & \checkmark & \checkmark & \underline{53.28} & \underline{51.01} & \underline{50.44} & \underline{50.06} & \underline{51.20} \\
\cmidrule{2-9}
& \multirow{3}{*}{WikiTQ} & \checkmark & $\times$ & \underline{43.05} & \underline{42.45} & 42.70 & \underline{42.96} & \underline{42.79} \\
&  & $\times$ & \checkmark & 42.77 & 41.55 & \underline{43.02} & 41.87 & 42.30 \\
&  & \checkmark & \checkmark & {\bf 43.09} & {\bf 42.63} & {\bf 43.39} & {\bf 43.21} & {\bf 43.08} \\
\cmidrule{2-9}
& \multirow{3}{*}{FeTaQA (BLEU)} & \checkmark & $\times$ & \underline{29.83} & 29.51 & 29.40 & 29.36 & 29.52 \\
&  & $\times$ & \checkmark & 29.69 & \underline{30.34} & {\bf 30.76} & {\bf 30.48} & {\bf 30.32} \\
&  & \checkmark & \checkmark & {\bf 30.20} & {\bf 30.55} & \underline{30.36} & \underline{29.96} & \underline{30.27} \\
\midrule
\multirow{9}{*}{Llama-3-8B-Instruct}
& \multirow{3}{*}{HiTab} & \checkmark & $\times$ & {\bf 72.98} & {\bf 74.56} & {\bf 73.36} & {\bf 75.13} & {\bf 74.01} \\
&  & $\times$ & \checkmark & 71.40 & 72.85 & 70.96 & 72.10 & 71.83 \\
&  & \checkmark & \checkmark & \underline{72.41} & \underline{73.80} & \underline{72.92} & \underline{74.43} & \underline{73.39} \\
\cmidrule{2-9}
& \multirow{3}{*}{WikiTQ} & \checkmark & $\times$ & \underline{59.62} & {\bf 60.04} & {\bf 59.83} & {\bf 59.74} & {\bf 59.81} \\
&  & $\times$ & \checkmark & {\bf 59.76} & 58.68 & 59.07 & 59.23 & 59.19 \\
&  & \checkmark & \checkmark & 59.42 & \underline{59.32} & \underline{59.46} & \underline{59.60} & \underline{59.45} \\
\cmidrule{2-9}
& \multirow{3}{*}{FeTaQA (BLEU)} & \checkmark & $\times$ & 32.31 & {\bf 33.18} & 32.70 & 32.32 & 32.63 \\
&  & $\times$ & \checkmark & {\bf 32.66} & 32.91 & \underline{32.71} & \underline{32.34} & \underline{32.66} \\
&  & \checkmark & \checkmark & \underline{32.39} & \underline{33.15} & {\bf 32.82} & {\bf 32.81} & {\bf 32.79} \\
\midrule
\multirow{9}{*}{Qwen2.5-7B}
& \multirow{3}{*}{HiTab} & \checkmark & $\times$ & \underline{64.90} & \underline{63.38} & {\bf 67.36} & {\bf 65.78} & \underline{65.36} \\
&  & $\times$ & \checkmark & 61.93 & 60.73 & 63.83 & 62.88 & 62.34 \\
&  & \checkmark & \checkmark & {\bf 66.92} & {\bf 65.78} & \underline{63.95} & \underline{65.59} & {\bf 65.56} \\
\cmidrule{2-9}
& \multirow{3}{*}{WikiTQ} & \checkmark & $\times$ & \underline{53.75} & {\bf 54.70} & \underline{53.34} & \underline{53.36} & \underline{53.79} \\
&  & $\times$ & \checkmark & 53.43 & 52.88 & 53.31 & 53.20 & 53.20 \\
&  & \checkmark & \checkmark & {\bf 54.33} & \underline{53.98} & {\bf 54.12} & {\bf 53.52} & {\bf 53.99} \\
\cmidrule{2-9}
& \multirow{3}{*}{FeTaQA (BLEU)} & \checkmark & $\times$ & {\bf 30.83} & {\bf 31.01} & \underline{30.96} & \underline{31.04} & \underline{30.96} \\
&  & $\times$ & \checkmark & 30.63 & \underline{30.97} & 30.70 & 30.78 & 30.77 \\
&  & \checkmark & \checkmark & \underline{30.82} & 30.83 & {\bf 31.08} & {\bf 31.15} & {\bf 30.97} \\
\bottomrule
\end{tabular}
\caption{\textbf{Per-seed accuracy (\%; BLEU for FeTaQA) of DeepTable variants on HiTab, WikiTQ, and FeTaQA under greedy decoding.}
All experiments use $\lambda_{\mathrm{SAB}}=\lambda_{\mathrm{TPE}}=1000$. Bold indicates the best result and underline the second-best, among the three DeepTable variants, per column.
}
\label{tab:app-hitab}
\end{table*}

\begin{table*}[t]
\centering\small
\begin{tabular}{lccccccc}
\toprule
 {\bf Backbone} & {\bf SAB} & {\bf TPE}& {\bf s0} & {\bf s1} & {\bf s2} & {\bf s3} &   {\bf Avg.} \\
\midrule
\multirow{3}{*}{%
\begin{tabular}{@{}l@{}}
DeepSeek-LLM-7B-Chat\\
\end{tabular}}
& \checkmark & $\times$   & {\bf 76.15} & {\bf 76.39} & {\bf 76.46} & {\bf 76.39} & {\bf 76.35} \\
&  $\times$ & \checkmark    & 74.65 & 74.48 & 74.51 & 74.12 & 74.44 \\
& \checkmark & \checkmark & 75.36 & 75.04 & 74.83 & 74.87 & 75.03 \\
\midrule
\multirow{3}{*}{Llama-3-8B-Instruct}
& \checkmark & $\times$   & {\bf 85.01} & {\bf 85.01} & {\bf 84.89} & {\bf 85.35} & {\bf 85.06} \\
& $\times$ & \checkmark    & 84.47 & 83.98 & 84.63 & 84.51 & 84.40 \\
& \checkmark & \checkmark & 84.90 & 82.75 & 84.83 & 84.94 & 84.36 \\
\midrule
\multirow{3}{*}{Qwen2.5-7B}
& \checkmark & $\times$   & {\bf 81.31} & {\bf 81.47} & {\bf 81.38} & 81.17 & {\bf 81.33} \\
& $\times$ & \checkmark     & 80.64 & 81.17 & 80.59 & 80.44 & 80.71 \\
& \checkmark & \checkmark & 80.87 & 81.39 & 81.05 & {\bf 81.23} & 81.14 \\
\bottomrule
\end{tabular}
\caption{Per-seed TabFact accuracy (\%) of DeepTable variants under greedy decoding, across all three LLM backbones.
}
\label{tab:app-tabfact}
\end{table*}

\section{Depth-Stratified Grid}
\label{app:depthgrid}
Table~\ref{tab:depthgrid} provides a finer-grained analysis of HiTab by jointly
stratifying samples according to the depths of the top and left header
hierarchies. For each depth combination, the table reports the four-seed average
accuracy of TableLoRA and DeepTable, together with the paired improvement
($\Delta$) and the corresponding sample count.

Consistent with the marginal analysis in Table~\ref{tab:depth}, DeepTable
improves upon TableLoRA for most depth combinations. The largest fluctuations
occur in the deepest buckets, which contain relatively few samples.

\begin{table}[t]
\centering
\small
\begin{tabular}{cc ccc}
\toprule
{\bf top} & {\bf left} & ${\mathbf{n}}$ & {\bf TableLoRA} & {\bf DeepTable} \\
\midrule
1 & 1 & 35  & 40.00 & 40.71~($+$0.71) \\
1 & 2 & 97  & 42.01 & 39.95~($-$2.06) \\
1 & 3 & 205 & 36.95 & 38.66~($+$1.71) \\
1 & 4 & 16  & 35.94 & 48.44~($+$12.50) \\
\midrule
2 & 1 & 88  & 38.64 & 41.76~($+$3.12) \\
2 & 2 & 516 & 54.60 & 57.03~($+$2.42) \\
2 & 3 & 307 & 36.32 & 39.50~($+$3.18) \\
2 & 4 & 14  & 23.21 & 17.86~($-$5.36) \\
\midrule
3 & 1 & 100 & 24.75 & 33.00~($+$8.25) \\
3 & 2 & 156 & 28.85 & 30.13~($+$1.28)\\
3 & 3 & 45  & 20.56 & 22.78~($+$2.22) \\
3 & 4 & 5   & 60.00 & 65.00~($+$5.00) \\
\bottomrule
\end{tabular}
\caption{Depth-stratified results on HiTab.
Samples are grouped by the depths of the top and left header hierarchies.
The table reports the four-seed average accuracy (\%) of TableLoRA and
DeepTable, together with the paired improvement ($\Delta$).
$n$ denotes the number of samples in each depth combination.}
\label{tab:depthgrid}
\end{table}

\section{TabFact True/False Asymmetry}
\label{app:tabfact_tf}

\begin{table}[t]
\centering
\setlength{\tabcolsep}{4.5pt}
\small
\begin{tabular}{llcccc}
\toprule
& &  {\bf SAB} & {\bf TPE}  & {\bf True} & {\bf False} \\
\midrule
\multicolumn{4}{l}{\textit{DeepSeek-LLM-7B-Chat}} \\
& TableLoRA  & $-$ & $-$ & 84.6  & 67.1   \\
& DeepTable & \checkmark & $\times$ & 84.1~($-$0.5) & {\bf 68.5}~($+$1.4)   \\
& DeepTable & $\times$ & \checkmark  & {\bf 87.4}~($+$2.8) & 61.3~($-$5.8)   \\
& DeepTable & \checkmark & \checkmark  & 86.8~($+$2.2) & 63.2~($-$3.9)   \\
\midrule
\multicolumn{4}{l}{\textit{Qwen2.5-7B}} \\
& TableLoRA & $-$ & $-$  & 86.6 & 75.7  \\
& DeepTable & \checkmark & $\times$  & 86.6~($+$0.0) & {\bf 76.0}~($+$0.3)   \\
& DeepTable & $\times$ & \checkmark  & 86.6~($+$0.0) & 74.7~($-$1.0)   \\
& DeepTable & \checkmark & \checkmark & 86.6~($+$0.0) & 75.6~($-$0.1)   \\
\bottomrule
\end{tabular}
\caption{TabFact four-seed accuracy (\%) by gold label.
Values in parentheses denote the differences from the corresponding
TableLoRA baseline.
The TabFact test set contains 6,425 True and 6,354 False examples.
}
\label{tab:tabfact_tf}
\end{table}

TabFact is a binary classification task, allowing the effects of the proposed
structural modules to be analyzed separately on True and False examples.
Table~\ref{tab:tabfact_tf} reports the four-seed average accuracy for each
gold label.

On DeepSeek-LLM-7B-Chat, TPE and \method\ exhibit different effects.
Compared with TableLoRA, TPE improves the accuracy on True examples by
2.8 percentage points but reduces the accuracy on False examples by
5.8 percentage points. In contrast, \method\ produces only a small change
on True examples ($-0.5$ points) while improving the accuracy on False
examples by 1.4 percentage points. DeepTable combines the two modules and
achieves intermediate results on both labels.

To further characterize this behavior, we computed the distribution of
predicted labels. Compared with TableLoRA, TPE increases the proportion
of predicted True labels from 58.9\% to 62.0\%, consistent with its higher
accuracy on True examples and lower accuracy on False examples.

\begin{table*}[t]
\centering\small
\setlength{\tabcolsep}{5pt}
\begin{tabular}{lll}
\toprule
& {\bf Configuration} & {\bf Value} \\
\midrule
\multirow{9}{*}{Model} & \multirow{3}{*}{Base model}    & DeepSeek-LLM-7B-Chat ($30$L, $32$H); \\
& & Qwen2.5-7B-Instruct ($28$L, $28$Q/$4$KV heads, GQA); \\
& & Llama-3-8B-Instruct ($32$L, $32$Q/$8$KV heads, GQA) \\
& LoRA / 2D-LoRA rank & $8$ \\
& Adapted modules         & key and value projections \\
& \multirow{3}{*}{\method\ parameters}     & DeepSeek:  $1{,}920$ scalars ($30{\times}32{\times}2$);\\
& &Qwen2.5: $1{,}568$ ($28{\times}28{\times}2$);\\
& &Llama-3: $2{,}048$ ($32{\times}32{\times}2$) \\
& TPE vocabulary       & $16{,}793$ nodes (HiTab) \\
\midrule
\multirow{8}{*}{Training} & Base learning rate      & $5{\times}10^{-6}$, cosine \\
& Add-on LR mult.\ $\lambda$ & $1000$ (shared across backbones) \\
& Epochs                  & $3$ (all datasets) \\
& Serialization cutoff    & $2000$ ($4000$ for TabFact) \\
& Effective batch size    & $16$ \\
& Precision               & bf16 \\
& Seeds                   & $0,1,2,3$ \\
& Total training compute  & ${\sim}3{,}041$ GPU-hours ($2{\times}$L40S) across $202$ runs \\
\midrule
\multirow{2}{*}{Inference}& Hardware                & $2\times$ NVIDIA L40S (48 GB) \\
& Decoding                & greedy \\

\bottomrule
\end{tabular}
\caption{Training hyperparameters used in all experiments.
The Qwen2.5-7B and Llama-3-8B-Instruct configurations differ from the
DeepSeek configuration only in the base model and the resulting number of
\method\ parameters.}
\label{tab:hparams}
\end{table*}

We further report the same analysis on Qwen2.5-7B. Compared with
DeepSeek-LLM-7B-Chat, the differences among the methods are considerably
smaller, although the reduction on False examples for TPE remains.

\section{Hyperparameters}
\label{app:hparams}
Table~\ref{tab:hparams} summarizes the training hyperparameters used in all
experiments. Unless otherwise specified, all settings follow
\citet{he2025tablelora}. The only modifications are those required by our
hardware (effective batch size and precision) and the learning-rate
multipliers introduced for \method\ and TPE.

The DeepSeek-LLM-7B-Chat and Qwen2.5-7B-Instruct experiments share the same
training configuration. The only differences are the base model itself and
the resulting number of \method\ parameters, which depends on the model
architecture.

\section{Implementation Details}
\label{app:impl}

\paragraph{Metadata pipeline.}
Each serialized example is accompanied by a per-cell metadata record,
including the row index, column index, cell text, top-header path,
left-header path, and cell type, where the cell type is one of
\{\textit{corner}, \textit{header-top}, \textit{header-left},
\textit{data}\}. During batching, the data collator assigns each cell's
row and column indices to all tokens belonging to that cell to construct
the attention masks for \method. Likewise, the corresponding top- and
left-header path-ID sequences are assigned to all tokens for TPE.
Variable-length header paths are padded to a common depth within each batch.

\paragraph{Attention implementation and load order.}
\method\ is implemented by patching both the eager and SDPA attention
implementations of the underlying LLMs. For DeepSeek-LLM-7B-Chat, we modify
\texttt{LlamaAttention} and \texttt{LlamaSdpaAttention}; for the
Qwen2.5-7B cross-model experiments, we modify
\texttt{Qwen2Attention} and \texttt{Qwen2SdpaAttention}. For grouped-query
attention, the structural bias is added to the attention logits after the
\texttt{repeat\_kv} expansion, with an independent learnable bias for each
query head.

In the eager implementation, the structural bias is added to the
pre-softmax attention logits immediately after applying the causal mask.
In the SDPA implementation, the same bias is incorporated into the
additive attention-mask argument. The attention patch must be installed
\emph{after} loading TableLoRA's prompt encoder; otherwise, the custom
data collator that provides the row and column indices is overwritten.
FlashAttention is disabled because it does not provide an interface for
injecting arbitrary additive attention biases, resulting in
approximately a 22\% reduction in training throughput.

\paragraph{Zero-initialization verification.}
As a sanity check, we evaluate a fixed validation batch using both
TableLoRA and the full \method $+$ TPE model before any optimization
(step~0). The two models produce bit-identical logits, confirming that
the additional parameters introduce no changes before training and that
all observed differences arise solely from subsequent parameter updates.

\end{document}